\documentclass[letterpaper]{article}
\usepackage{aaai2027}
\nocopyright

\usepackage{times}
\usepackage{helvet}
\usepackage{courier}
\usepackage[hyphens]{url}
\usepackage{graphicx}
\usepackage{natbib}
\usepackage{caption}
\usepackage{booktabs}
\usepackage{amsmath}
\usepackage{tikz}
\usetikzlibrary{arrows.meta,positioning,calc,fit,backgrounds,shapes.callouts,shapes.symbols}
\definecolor{axblue}{RGB}{59,110,165}
\definecolor{axgreen}{RGB}{74,124,89}
\definecolor{axamber}{RGB}{224,160,77}
\definecolor{axpurple}{RGB}{123,108,168}

\newcommand{\zhi}[1]{\raisebox{-0.2em}{\includegraphics[height=1.0em]{zh/#1}}}
\newcommand{\icn}[2][0.95em]{\raisebox{-0.15em}{\includegraphics[height=#1]{figures/icons/pdf/#2}}}

\ifdefined{ /TemplateVersion (2027.1) }\fi
\title{Beyond Similarity: Grounded Agentic Extraction and Expert-Adjudicated\\
Evaluation of Intertextuality in Classical Chinese Histories}
\author{Zhaoji Wang\textsuperscript{\rm 1,2}, Wanyu Si\textsuperscript{\rm 1,2}, Jun Wang\textsuperscript{\rm 1,2,*}}
\affiliations{\textsuperscript{\rm 1}Department of Information Management, Peking University, Beijing, China\\
\textsuperscript{\rm 2}Research Center for Digital Humanities, Peking University, Beijing, China\\
zhaojiwang@stu.pku.edu.cn, 2401211838@stu.pku.edu.cn, junwang@pku.edu.cn\\
\textsuperscript{\rm *}Corresponding author.}

\begin{document}
\maketitle

\begin{abstract}
Computational approaches to intertextuality have advanced from string matching
to neural retrieval, yet their outputs, similarity scores and parallel-passage
lists, identify where texts reuse one another without characterizing \emph{how}
or \emph{why}. We recast fine-grained intertextuality extraction as an
agentic task in which a large language model (LLM) reads two text units in full and,
through a constrained tool interface, must ground each proposed reuse in exact
character spans on both sides and label it under a five-dimension
typology of reuse (form, aspect, source-marking, function, stance). We validate the
approach on an exhaustive comparison of the \emph{Analects} with the \emph{Book
of Han}, where three domain experts adjudicate a pooled multi-model candidate set
into a benchmark of 2{,}533 intertextual pairs. Against this standard we study twelve LLMs,
reporting precision (56\%--93\%), a 51$\times$ cost spread at comparable quality,
and how well their confidence is calibrated. Expert agreement traces a
\emph{reliability gradient}: dimensions legible on the textual surface are
annotated consistently, while those requiring inference of intent are contested,
delimiting the claims such annotation supports. Scaling the validated extractor to the full
Twenty-Four Histories (65{,}380 comparisons, 5{,}766 pairs) recovers corpus-level
structure a similarity score cannot express. The interpretive composition of
citation shows no systematic change across eighteen centuries, yet the same
passage is quoted
less and less literally. Stability in the aggregate with drift in the
individual case is what a cultural-attraction account expects. We release the
extraction protocol and the expert-adjudicated benchmark.
\end{abstract}

\section{Introduction}
In the \emph{Book of Han} (first century~CE), the
abdication charge from the \emph{Analects},
``\zhi{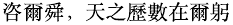}'' (the heaven-ordained
succession rests upon your person), is transcribed almost verbatim. A few
chapters later, in the biography of the usurper Wang Mang, the same line reappears
as ``\zhi{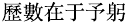}'', the second-person pronoun
quietly rewritten into the first. By the 260s, in the abdication
documents of the \emph{Records of the Three Kingdoms}, only a formulaic skeleton
survives, ``\zhi{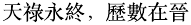}'', detached from the
\emph{Analects} and fused into the standard script of dynastic transfer. One
canonical sentence, three degrees of fidelity, spread across three centuries of
official historiography.

\begin{figure*}[t]
\centering
\resizebox{\textwidth}{!}{%
\begin{tikzpicture}[
  font=\small, >=Stealth,
  src/.style={draw=gray!60,rounded corners=2pt,fill=gray!7,align=center,inner sep=4pt,font=\footnotesize,minimum width=50mm},
  agentbox/.style={draw=axblue!85,thick,rounded corners=3pt,fill=axblue!7,align=center,inner sep=5pt,font=\footnotesize},
  tool/.style={draw=axgreen!85,rounded corners=2pt,fill=axgreen!8,align=center,inner sep=4pt,font=\footnotesize},
  chk/.style={draw=axblue!60,rounded corners=2pt,fill=white,align=center,inner sep=4pt,font=\footnotesize},
  gate/.style={draw=axblue!60,rounded corners=1.5pt,fill=white,align=left,inner sep=3pt,font=\scriptsize,text width=40.5mm},
  pairc/.style={draw=axamber!95!black,thick,rounded corners=3pt,fill=axamber!14,align=left,inner sep=5pt,font=\footnotesize},
  stat/.style={draw=gray!55,rounded corners=1.5pt,fill=gray!4,align=center,inner sep=3.5pt,font=\scriptsize},
]
\node[src,anchor=west] (a) at (0,0) {\icn{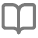}\, chunk $A$ (\emph{Analects} book)\\[1pt]\ldots\ ``\zhi{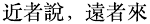}''\ \ldots};
\node[src,anchor=north west] (b) at ($(a.south west)+(0,-7mm)$) {\icn{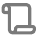}\, chunk $B$ (history scroll)\\[1pt]\ldots\ ``\zhi{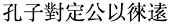}''\ \ldots};
\node[agentbox,right=10mm of $(a.east)!0.5!(b.east)$] (agent) {\icn{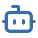}\, LLM agent\\reads both chunks in full,\\proposes candidate reuses\\[1.5pt]
  \icn[0.8em]{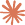}\,\icn[0.8em]{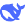}\,\icn[0.8em]{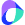}\,\icn[0.8em]{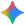}\,\icn[0.8em]{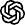}\,\icn[0.8em]{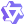}\,\icn[0.8em]{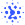}};
\node[tool,above=7mm of agent,minimum width=52mm] (pos) {\icn{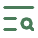}\, \texttt{find\_position}\\exact search $\to$ $[\mathit{start},\mathit{end})$ + context};
\node[font=\footnotesize\bfseries,anchor=north west,align=left] (heada) at ($(a.west |- pos.north)+(-0.8mm,0)$) {(a) one extraction task\\over one chunk pair};
\coordinate (cheadx) at ($(agent.east)+(15.5mm,0)$);
\coordinate (cheady) at ($(pos.north)+(0,-2mm)$);
\node[font=\scriptsize,anchor=north west,text=axblue!60!black]
  (chead) at (cheadx |- cheady)
  {\icn{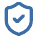}\, \texttt{add\_pair}: 3 checks};
\node[gate,anchor=north west] (g1) at ($(chead.south west)+(0,-2.0mm)$)
  {\icn{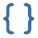}\, \textbf{schema}: 5-dim labels $\cdot$ spans\\$\cdot$ confidence};
\node[gate,anchor=north west] (g2) at ($(g1.south west)+(0,-3.4mm)$)
  {\icn{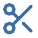}\, \textbf{verbatim re-slice}: chunk$[\mathit{start},\mathit{end})$\\$=$ fragment, char-for-char};
\node[gate,anchor=north west] (g3) at ($(g2.south west)+(0,-3.4mm)$)
  {\icn{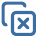}\, \textbf{six-tuple dedup}: the same\\intertextual pair, keyed by\\($\mathit{text}_A$, $\mathit{start}_A$, $\mathit{end}_A$,\\$\mathit{text}_B$, $\mathit{start}_B$, $\mathit{end}_B$),\\cannot enter twice};
\node[font=\tiny,text=red!55!black,anchor=north west] (cfoot) at ($(g3.south west)+(0,-2.8mm)$)
  {any check fails $\Rightarrow$ pair rejected, error returned};
\begin{scope}[on background layer]
\node[draw=axblue!60,rounded corners=3pt,fill=axblue!3,inner sep=2mm,
      fit=(chead)(g1)(g2)(g3)(cfoot)] (commit) {};
\end{scope}
\draw[->,axblue!60!black] (g1) -- node[right,font=\tiny]{pass} (g2);
\draw[->,axblue!60!black] (g2) -- node[right,font=\tiny]{pass} (g3);
\node[tool,minimum width=52mm,anchor=south] (aux) at (agent.south |- commit.south)
  {\icn{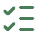}\, \texttt{list}\,/\,\texttt{remove}\\inspect \& retract committed\\intertextual pairs};
\node[font=\scriptsize,text=gray!25!black,align=left,anchor=south west,inner sep=1pt]
  (ctxnote) at (a.west |- commit.south)
  {both chunks enter the model's context \emph{whole}:\\
   one \emph{Analects} book, one full history scroll.\\
   No sentence splitting, no embedding retrieval---\\
   no similarity filter decides what the model sees.};
\node[pairc,anchor=west] (pair) at ($(commit.east |- agent.east)+(11mm,3.5mm)$)
  {\icn{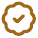}\, \textbf{committed intertextual pair}\\[0.5pt]
{\scriptsize\color{gray!40!black}\emph{Analects} Book 13 $\to$ \emph{Book of Han} scroll 6}\\[1pt]
$A\,[876,883)$:\ ``\zhi{inva}''\\
$B\,[3156,3164)$:\ ``\zhi{invb}''\\[2.5pt]
{\scriptsize\icn{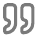}\,{\color{gray!45!black}form:} {\color{axpurple!85!black}paraphrase} \ \icn{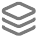}\,{\color{gray!45!black}aspect:} {\color{axpurple!85!black}content}}\\[1pt]
{\scriptsize\icn{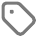}\,{\color{gray!45!black}marking:} {\color{axpurple!85!black}marked} \ \icn{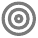}\,{\color{gray!45!black}function:} {\color{axpurple!85!black}application} \ \icn{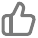}\,{\color{gray!45!black}stance:} {\color{axpurple!85!black}positive}}\\[1pt]
{\scriptsize\icn{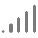}\,{\color{gray!45!black}evidence:} {\color{axpurple!85!black}medium} \ \icn{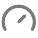}\,{\color{gray!45!black}confidence:} {\color{axpurple!85!black}0.72}}};
\begin{scope}[on background layer]
\draw[draw=axamber!75!black!60,fill=axamber!8,rounded corners=3pt]
  ($(pair.south west)+(2.2mm,-2.2mm)$) rectangle ($(pair.north east)+(2.2mm,-2.2mm)$);
\draw[draw=axamber!75!black!60,fill=axamber!10,rounded corners=3pt]
  ($(pair.south west)+(1.1mm,-1.1mm)$) rectangle ($(pair.north east)+(1.1mm,-1.1mm)$);
\end{scope}
\coordinate (pairshadow) at ($(pair.south east)+(2.2mm,-2.2mm)$);
\node[font=\footnotesize\bfseries,anchor=north west] (headb) at ($(pair.west |- pos.north)+(-0.8mm,0)$) {(b) verifiable outputs};
\node[align=left,font=\scriptsize,text=gray!25!black,anchor=south west,inner sep=1pt]
  (simview) at (pair.west |- commit.south)
  {The two fragments share zero characters: string similarity scores\\
   this real intertextual pair 0.00, and dense retrieval ranks its\\
   source \#232 of 1{,}470---only reading recovers \emph{how} and \emph{why}.};
\begin{scope}[on background layer]
\node[draw=gray!45,rounded corners=4pt,inner sep=2mm,
      fit=(heada)(headb)(a)(b)(ctxnote)(agent)(pos)(aux)(commit)(pair)(pairshadow)(simview)] (bigbox) {};
\end{scope}
\coordinate (ctop) at ($(bigbox.south)+(0,-2.8mm)$);
\node[font=\footnotesize\bfseries,anchor=north west] at ($(bigbox.west |- ctop)+(1.2mm,-1.2mm)$) {(c) task termination};
\tikzset{
  cframe/.style={draw=gray!50,line width=0.5pt,rounded corners=2.5pt,fill=gray!3},
  cnum/.style={circle,draw=gray!55,fill=white,font=\tiny\bfseries,text=gray!30!black,inner sep=0.5pt},
  ccap/.style={font=\scriptsize,text=gray!25!black,align=center,anchor=south},
}
\def\fw{74.4mm}\def\gut{5.5mm}\def\fh{34mm}
\node[cframe,minimum width=\fw,minimum height=\fh,anchor=north west]
  (f1) at ($(bigbox.west |- ctop)+(2.2mm,-6.6mm)$) {};
\node[cframe,minimum width=\fw,minimum height=\fh,anchor=north west]
  (f2) at ($(f1.north east)+(\gut,0)$) {};
\node[cframe,minimum width=\fw,minimum height=\fh,anchor=north west]
  (f3) at ($(f2.north east)+(\gut,0)$) {};
\node[cnum,anchor=north west] at ($(f1.north west)+(1mm,-1mm)$) {1};
\node[cnum,anchor=north west] at ($(f2.north west)+(1mm,-1mm)$) {2};
\node[cnum,anchor=north west] at ($(f3.north west)+(1mm,-1mm)$) {3};
\node (bot1) at ($(f1.center)+(-11.5mm,-3mm)$) {\icn[9mm]{bot_blue}};
\foreach \d in {2.0,1.0,0}{%
  \draw[draw=axamber!80!black,fill=axamber!13,rounded corners=0.6pt]
    ($(bot1.east)+(6.5mm,-4mm)+(\d mm,\d mm)$) rectangle ++(8.5mm,11mm);}
\node[font=\scriptsize,anchor=west] (cardlab) at ($(bot1.east)+(18mm,2mm)$) {$\times\,n$};
\node[font=\tiny,text=gray!55!black,anchor=north] at ($(cardlab.south)+(0,-0.4mm)$) {$(n\!\geq\!0)$};
\node[cloud,cloud puffs=11,cloud puff arc=110,aspect=2.5,draw=gray!55,fill=white,
      font=\tiny,align=center,inner sep=0pt,minimum width=26mm,minimum height=8.5mm]
  (th1) at ($(bot1.north)+(6.5mm,8.5mm)$) {nothing left\\to add?};
\fill[white,draw=gray!55] ($(bot1.north)+(3mm,2.6mm)$) circle (0.9mm);
\fill[white,draw=gray!55] ($(bot1.north)+(4.6mm,4.4mm)$) circle (0.55mm);
\node[ccap] at ($(f1.south)+(0,1.8mm)$) {\textbf{a run ends}: $n$ committed pairs};
\node (bot2) at ($(f2.center)+(-13.5mm,-2mm)$) {\icn[9mm]{bot_blue}};
\coordinate (gx) at ($(bot2.east)+(9mm,0mm)$);
\fill[gray!45] ($(gx)+(-0.7mm,-6.5mm)$) rectangle ++(1.4mm,13mm);
\fill[gray!45] ($(gx)+(6.3mm,-6.5mm)$) rectangle ++(1.4mm,13mm);
\foreach \i in {0,1,2,3}{%
  \fill[red!55!black] ($(gx)+(0,\i*2.8mm-4.9mm)$) rectangle ++(7mm,1.4mm);}
\node at ($(gx)+(-2.6mm,3mm)$) {\icn[6mm]{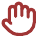}};
\draw[->,red!60!black,line width=0.6pt,rounded corners=0.7mm]
  ($(bot2.east)+(1.2mm,-1.6mm)$) -- ($(gx)+(-1.4mm,-1.6mm)$) --
  ($(gx)+(-1.4mm,-4.4mm)$) -- ($(bot2.east)+(1.2mm,-4.4mm)$);
\node[ellipse callout,callout absolute pointer={($(gx)+(1mm,3mm)$)},callout pointer width=3mm,
      draw=red!55!black,fill=red!3,text=red!55!black,font=\tiny,align=center,inner sep=1.6pt]
  (sb2) at ($(gx)+(12mm,8.5mm)$) {call \texttt{submit}\\first};
\node[ccap] at ($(f2.south)+(0,1.8mm)$) {\textbf{stop guard}: no exit but \texttt{submit}};
\node (bot3) at ($(f3.center)+(-17.5mm,1.25mm)$) {\icn[9mm]{bot_blue}};
\node (snd) at ($(bot3.east)+(6.5mm,0)$) {\icn[6.5mm]{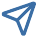}};
\node[font=\tiny,text=axblue!60!black,align=center,anchor=south] at ($(snd.north)+(0,0.8mm)$) {status\\check};
\def\doorw{6mm}\def\doorh{8.5mm}
\coordinate (mdoor) at ($(f3.center)+(6mm,3mm)$);
\draw[draw=axgreen!70,fill=axgreen!10,rounded corners=1pt] (mdoor) rectangle ++(\doorw,\doorh);
\fill[axgreen!70] ($(mdoor)+(0.82*\doorw,0.45*\doorh)$) circle (0.5mm);
\node at ($(mdoor)+(0.42*\doorw,0.52*\doorh)$) {\icn[3.6mm]{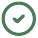}};
\node[font=\tiny,anchor=west,align=left] at ($(mdoor)+(\doorw+0.9mm,0.5*\doorh)$)
  {{\color{axgreen!60!black}\textbf{match}}\\{\color{gray!55!black}$(n\!\geq\!1)$}};
\coordinate (ndoor) at ($(f3.center)+(6mm,-9mm)$);
\draw[draw=gray!55,fill=gray!6,rounded corners=1pt] (ndoor) rectangle ++(\doorw,\doorh);
\fill[gray!55] ($(ndoor)+(0.82*\doorw,0.45*\doorh)$) circle (0.5mm);
\node at ($(ndoor)+(0.42*\doorw,0.52*\doorh)$) {\icn[3.6mm]{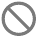}};
\node[font=\tiny,anchor=west,align=left] at ($(ndoor)+(\doorw+0.9mm,0.5*\doorh)$)
  {{\color{gray!35!black}\texttt{no\_match}}\\{\color{gray!55!black}$(n\!=\!0)$}};
\draw[->,axgreen!70!black] (snd.east) -- ($(mdoor)+(0,0.5*\doorh)$);
\draw[->,gray!55!black]    (snd.east) -- ($(ndoor)+(0,0.5*\doorh)$);
\node[font=\tiny,text=red!55!black,align=center,anchor=north] (failn)
  at ($(snd.south)+(0,-4.2mm)$) {\icn[0.75em]{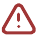}\,any other end: \emph{failed}};
\draw[->,red!55!black,dashed,line width=0.4pt]
  ($(snd)+(0,-2.8mm)$) -- ($(failn.north)+(0,0.3mm)$);
\node[ccap] at ($(f3.south)+(0,1.8mm)$) {\texttt{submit} \textbf{decides} the outcome};
\coordinate (cbot) at ($(f1.south)+(0,-2.6mm)$);
\begin{scope}[on background layer]
\draw[gray!45,rounded corners=4pt] (bigbox.west |- ctop) rectangle (bigbox.east |- cbot);
\end{scope}
\foreach \n in {pos,aux,commit}
  \node[font=\tiny,text=gray!55!black,anchor=north east] at ($(\n.north east)+(-1mm,-0.6mm)$) {tool};
\draw[->] (a.east) -- ([yshift=3.5mm]agent.west);
\draw[->] (b.east) -- ([yshift=-3.5mm]agent.west);
\draw[->,axgreen!70!black] ([xshift=-3.5mm]pos.south |- agent.north)
  -- node[left,font=\scriptsize]{locate} ([xshift=-3.5mm]pos.south);
\draw[->,axgreen!70!black,dashed] ([xshift=3.5mm]pos.south)
  -- node[right,font=\scriptsize]{offsets} ([xshift=3.5mm]pos.south |- agent.north);
\draw[->,axgreen!70!black] ([xshift=-3.5mm]aux.north |- agent.south)
  -- node[left,font=\scriptsize]{revise} ([xshift=-3.5mm]aux.north);
\draw[->,axgreen!70!black,dashed] ([xshift=3.5mm]aux.north)
  -- node[right,font=\scriptsize]{state} ([xshift=3.5mm]aux.north |- agent.south);
\draw[->] ([yshift=3.5mm]agent.east)
  -- node[above,font=\scriptsize]{propose} ([yshift=3.5mm]agent.east -| commit.west);
\draw[->,dashed,red!55!black] ([yshift=-4mm]agent.east -| commit.west)
  -- node[below,font=\scriptsize]{reject} ([yshift=-4mm]agent.east);
\draw[->] (pair.west -| commit.east) -- node[above,font=\scriptsize]{accept} (pair.west);
\draw[->,gray!55!black] (f1.east) -- (f2.west);
\draw[->,gray!55!black] (f2.east) -- (f3.west);
\def\statgap{16.75mm}
\coordinate (dtop) at ($(cbot)+(0,-2.8mm)$);
\node[font=\footnotesize\bfseries,anchor=north west] at ($(bigbox.west |- dtop)+(1.2mm,-1.2mm)$) {(d) validate-then-scale};
\node[stat,anchor=north west] (s1) at ($(bigbox.west |- dtop)+(2.2mm,-6.6mm)$)
  {\icn{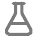}\, \textbf{validate}: 2{,}400 chunk-pair\\tasks $\times$ 12 LLMs};
\node[stat,anchor=east] (s5) at ($(bigbox.east |- s1.east)+(-2.2mm,0)$)
  {\icn{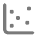}\, \textbf{scale}: 24 histories, 65{,}380 tasks\\$\to$ 5{,}766 intertextual pairs $\to$ clusters};
\node[stat,anchor=west] (s2) at ($(s1.east)+(\statgap,0)$) {6{,}018 proposed intertextual pairs\\$\to$ 3{,}489 pooled candidate pairs};
\node[stat,anchor=east] (s4) at ($(s5.west)+(-\statgap,0)$) {\icn{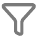}\, select extractor:\\precision $\cdot$ cost $\cdot$ calibration};
\node[stat] (s3) at ($(s2.east)!0.5!(s4.west)$) {\icn{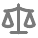}\, 2 experts + arbiter\\$\to$ 2{,}533 gold intertextual pairs};
\draw[->] (s1) -- (s2); \draw[->] (s2) -- (s3);
\draw[->] (s3) -- (s4); \draw[->] (s4) -- (s5);
\coordinate (dbot) at ($(s1.south)+(0,-2.2mm)$);
\begin{scope}[on background layer]
\draw[gray!45,rounded corners=4pt] (bigbox.west |- dtop) rectangle (bigbox.east |- dbot);
\end{scope}
\end{tikzpicture}}
\caption{Framework overview. (a)~The extraction agent for one chunk pair and
its task tools (names abbreviated): \texttt{find\_position}
grounds candidate wordings in exact offsets; \texttt{list}\,/\,\texttt{remove}
support revision; \texttt{add\_pair} commits a pair only past the three checks
shown. (b)~The
committed intertextual pair shown is real and shares zero characters across its fragments:
an edict of Emperor Wu invoking \emph{Analects} 13 by the sage's name alone
(cf.\ Beyond overlap).
(c)~A task can end only through \texttt{submit} (a stop guard blocks any
earlier exit), and \texttt{submit} itself checks the declared status against
the committed set; any other end is recorded as \emph{failed}.
(d)~The validate-then-scale pipeline.}
\label{fig:agent}
\end{figure*}
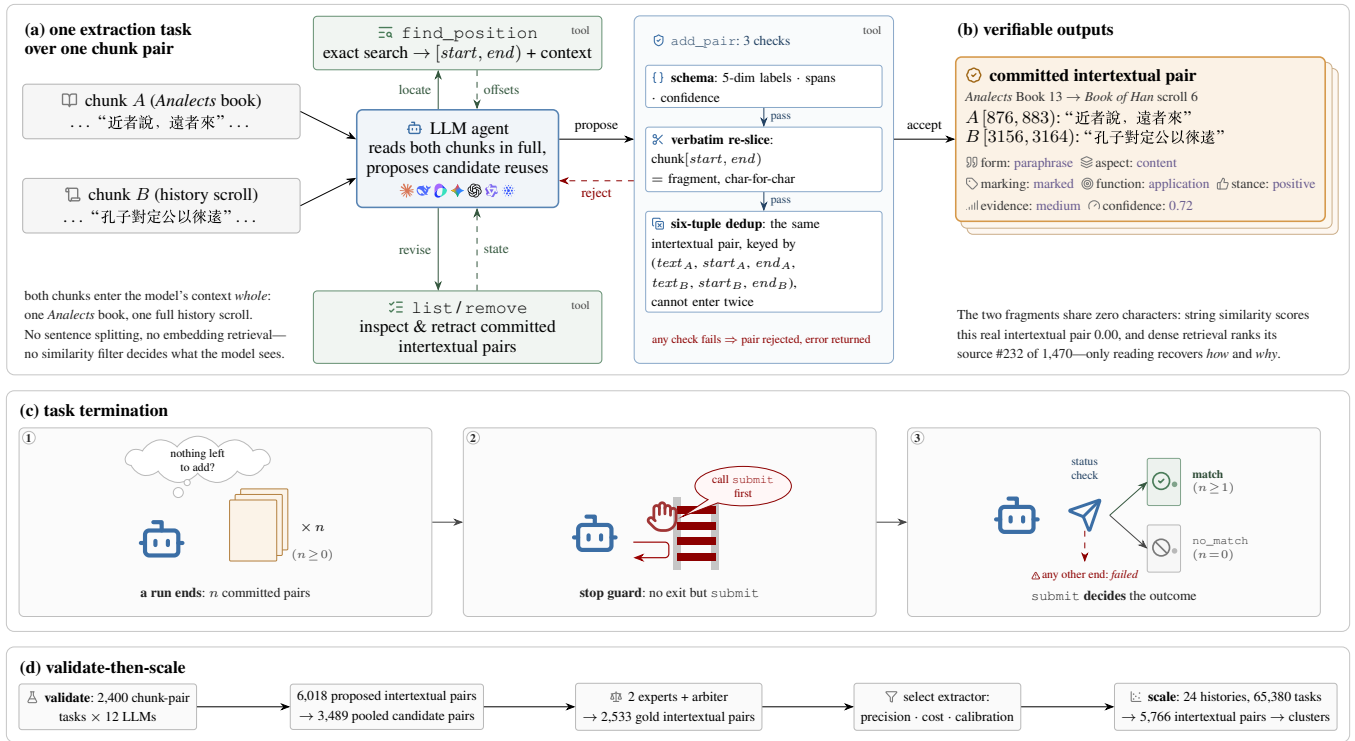

A text-reuse detector can find that these passages resemble the \emph{Analects};
it cannot say that the first is a faithful citation, the second an appropriation
of authority, the third a dead formula. This is the gap we address, and it is
consequential: a similarity score answers none of the questions historians
ask of this corpus. Over roughly
two millennia, China's official histories were compiled by authors steeped in the
classics, who quoted, paraphrased, and silently absorbed the \emph{Analects} into
judgments, memorials, and narrative alike. Tracing that reuse by hand is the
labor of concordance scholarship, and it has carried arguments of the first
rank. Gu Jiegang dated the strata of the legendary past by setting text
against text. The \emph{Odes} know Yu but not Yao and Shun. The
\emph{Analects}, later, praises both. The chapter that first chains them
into a succession, the source of this paper's opening abdication charge, he
read as a late imitation of archaic style. Each age, he argued, extended
the history it received, the thesis that launched the Doubting Antiquity
movement \citep{gu1923gushibian}. Comparison of this kind is bounded by
sustained expert attention, and in classical studies that attention is
scarce. Gu himself planned to proceed one book at a time. It does not
scale. Nor do the standard
computational substitutes. On our data, retaining every human-confirmed reuse
lets dense and sparse embedding prefilters discard only 2.5\% and 1.1\% of
chunk pairs (0.3\% combined), because literal and distributional similarity
systematically miss paraphrase, precisely the cases that matter. The
limit is representational: routing every comparison
through one similarity value keeps only the degree of reuse and discards its
kind.

We therefore formalize \emph{fine-grained intertextuality extraction} as
span-grounded relation extraction between two texts. Given a pair of text
units, a model must locate every
reused fragment on both sides as an exact character span, and label each resulting
\emph{intertextual pair} under a five-dimension typology of reuse. This is a localization-and-classification task with verifiable
outputs, not a similarity score. A prediction is correct only if the fragment
texts and their in-unit character offsets match. We instantiate it as an
\emph{agentic} task in which an LLM, constrained by a tool interface, commits only
span-verified, schema-valid pairs.

Because exhaustive expert judgment is affordable only at small scale, we adopt a
\emph{validate-then-scale} design. We validate on the \emph{Analects} against the
\emph{Book of Han}, where three domain experts adjudicate a pooled candidate set
to a gold standard and twelve LLMs are measured against it. We then apply the
validated extractor to the full Twenty-Four Histories and read its output as a
distribution of calibrated model judgments.

Our contributions are: (1)~a verifiable extraction protocol for fine-grained
intertextuality, executed by a tool-constrained LLM agent (Fig.~\ref{fig:agent}), where annotations exist
only as commitments that are span-verified, schema-checked, and de-duplicated
at write time, abstention is explicit, and execution is reproducible and
hermetic, so corpus-scale output is auditable rather than a parse of model
prose; (2)~an expert-adjudicated benchmark and a
twelve-model study, with 2{,}533 gold pairs adjudicated from 3{,}489 triple-annotated
candidates over the \emph{Analects}--\emph{Book of Han}, used to compare twelve
LLMs on precision, cost, and calibration, and to surface a reliability gradient
in expert agreement across the five dimensions; and (3)~a validated large-scale
application to all Twenty-Four Histories (65{,}380 comparisons, 5{,}766 pairs)
that recovers structure a similarity score cannot express, with the interpretive
composition of \emph{Analects} citation showing no systematic change across
eighteen centuries
while literal fidelity to a fixed passage declines.

\section{Related Work}
\paragraph{Text reuse across traditions.}
Computational study of intertextuality began with surface matching and sequence
alignment, pursued largely within separate philological traditions
\citep[for surveys, see][]{duan2025survey,sommerschield2023survey}. An early
computational model characterized reuse in the Greek New Testament
\citep{lee2007reuse}, and
for Latin poetry the Tesserae project matches shared words and lemmata to surface
parallels \citep{coffee2013tesserae}. Alignment-based systems detect reprinted
passages in nineteenth-century newspapers \citep{smith2013infectious}, parallel
passages in Hebrew--Aramaic and Buddhist Chinese corpora
\citep{shmidman2018parallel,nehrdich2020buddhist}, related verses in the Qur'an
\citep{sharaf2012qursim}, and reuse in Chinese corpora
with a language-agnostic aligner \citep{vierthaler2019blast}. For premodern
Chinese, \citet{sturgeon2018unsupervised} detects text reuse across the
transmitted early corpus, the Evol line scales sentence-embedding retrieval to
millions of intertextual pairs and mines the resulting networks for
cultural-evolution questions \citep{duan2023disentangling,wang2024evol}, and,
closest to us in corpus, \citet{deng2022shiji} compare the \emph{Shiji} and
\emph{Book of Han} through intertextual pairs. These methods scale, but operate
on literal or distributional overlap, return passage-level candidates, and
assign no typology of how a source is reused. Adding distributional semantics
widens coverage, monolingually and across languages
\citep{scheirer2016sense,manjavacas2019feasibility,burns2021profiling,riemenschneider2023graecia},
without changing this picture.

\paragraph{From detection to a typology of reuse.}
That reuse comes in kinds is recognized \citep{forstall2019quantitative}, and a
digital-humanities strand argues for moving ``from quantitative to qualitative
analysis'' and toward a computational hermeneutics of \emph{how}, not \emph{how
much}, texts are reused \citep{roe2024textreuse,moritz2016nonliteral}. LLMs have
very recently been turned on intertextuality. At NLP venues,
\citet{yang2025interideas} use LLMs to build a dataset of intertextual relations
among philosophical texts, and \citet{periti2024trotr} benchmark how already-known
reused passages are recontextualized. Both operate on relations between texts or
given reuse pairs, not on extracting and localizing reuse. Extraction itself
remains in preprints and digital-humanities venues, and none of these efforts
combine the pieces we do. \emph{Loci Similes}
\citep{schelb2026loci} pairs a directed corpus with a two-way typology and
expert-checked gold, but operates at the segment level and localizes no
character spans.
\citet{lau2024mining} mine asymmetric intertextuality from large Chinese corpora
with an LLM-assisted pipeline framed as retrieval, with no reuse typology. A
retrieval framing is moreover bounded by its
segmentation. Where several passages on one side answer to one on the other,
segment-to-segment matching cannot represent the relation, a structure our
corpus-scale run
surfaces below. \citet{umphrey2024expert} prompt an LLM to emit
intertextual pairs for expert validation, without span localization or
adjudicated gold. \citet{cameron2026horse} traces biblical allusion in a modern
novel. No prior system, in any tradition, combines generative-LLM extraction,
two-sided character-span localization, a multi-dimension reuse typology, and an
expert-adjudicated gold standard with reported agreement, applied at corpus scale.
We target that combination, and validate it before applying it at scale. A further
difference is architectural: these systems read conclusions out of model prose,
whereas our annotations exist only as tool commitments checked at write time,
which is what makes corpus-scale output auditable rather than parsed.

\section{Task and Agentic Extraction Protocol}
\paragraph{Task.}
We segment the two sources into text units (\emph{chunks}): the \emph{Analects}
into its 20 books, each history into its scrolls. Given a chunk pair $(A,B)$, the
task is to output every \emph{intertextual pair}: a fragment of $A$ and a fragment
of $B$, each as an exact character span $[\mathit{start},\mathit{end})$, together
with a label along five interpretive dimensions: \emph{form} (direct quotation
vs.\ paraphrase), \emph{aspect} (content vs.\ structure), \emph{source-marking}
(explicit, marked, unmarked), \emph{function} (background, support, application,
critique), and \emph{stance} (positive, neutral, negative, complex). Intertextual
reuse is defined strictly as \emph{traceable textual dependence}: shared topic,
agreement, common vocabulary, or diffuse influence are excluded. A prediction is
correct only if both fragment texts occur verbatim at their stated offsets.
Identity is likewise strict: two candidates are the same only if they agree on
the chunk pair, both fragment texts, and both offset ranges. Endpoint punctuation
differences and shifted span boundaries are distinct candidates. Precision is
the share of committed pairs judged valid against the adjudicated gold standard.

We cast this as an agentic task in the tool-using paradigm
\citep{yao2023react,schick2023toolformer,wang2024agentsurvey}. For each chunk pair,
an LLM agent reads both texts in full and commits candidate pairs through a
constrained tool interface. The contribution is the verifiability contract
this interface enforces and what that contract makes measurable. Both chunks enter the context whole, a complete
book against a complete scroll, rather than cut into sentences and paired by
$n$-gram or embedding similarity
\citep{sturgeon2018unsupervised,duan2023disentangling,wang2024evol}. An index
surfaces only what it already scores as close (the bottleneck quantified
above), and marking and rhetorical function are legible only in discourse that
sentence-level pairing discards.
Three
design principles address the failure modes of using an LLM for this task. The
concrete tool schemas, isolation settings, and control parameters are given in the
supplementary material.

\paragraph{P1: Verifiable grounding.}
An LLM asked for the exact reused spans on both sides produces fluent but
sometimes unanchored or fabricated parallels. The agent must therefore ground every proposed pair in exact
character offsets, obtained from a positioning tool, before it can be recorded.
The positioning tool is exact substring search over a chunk, returning
$[\mathit{start},\mathit{end})$ offsets with $\pm$30 characters of surrounding
context, queried singly or in batch. At commit time, the protocol re-slices each
chunk at the stated offsets and rejects any pair whose cited fragment does not
reproduce verbatim. This turns plausible-looking output into checkable textual
evidence, echoing the grounded-parallel emphasis of human-in-the-loop systems
for ancient texts \citep{assael2025aeneas}. In the twelve-model run below,
agents issued 319{,}465 positioning calls (about 11 per task attempt), and the
gates rejected 39\% of \texttt{add\_pair} calls, a quarter at the verbatim
re-slice, another 15\% at the schema check. Rejection tracks grounding, not
quality. The lowest-precision model alone drew two-thirds of the re-slice
rejections, yet 81\% of attempts that hit any rejection still ended with an
accepted pair. Ablating the positioning tool multiplies the gate-rejection rate
ninefold and output tokens fivefold (supplementary ablation).

\paragraph{P2: Tool-mediated structured commitment.}
Rather than parse conclusions from the model's prose, the agent commits each
annotation through a submission
interface that enforces the schema and rejects duplicate pairs at commit time,
duplicate identity being the six-tuple of both fragment texts and their offsets.
Six task tools implement the protocol: single and batch position lookup, pair
commitment, listing and removal of committed pairs (the only route to revising an
accepted pair), and a final submission call. A task's output is exactly the set of
committed pairs. The model's final free text is never parsed. Execution follows
the same discipline: each task runs hermetically, with a fresh temporary home and no
inherited configuration, and is keyed by a content hash of every input that determines
its result, including provider identity and endpoint, so the same model name
served by two providers cannot silently reuse cached results. Configuration
drift between runs is an error.

\paragraph{P3: Enforced completion and explicit abstention.}
At corpus scale every task must end in a well-defined outcome so that aggregate
statistics are trustworthy. Every task is classified
into exactly one of three outcomes: the agent commits at least one grounded
pair (a match), explicitly declares that the
two chunks share no reuse (abstention), or is recorded as a failure.
Concretely, a stop guard blocks the agent from
terminating before it has submitted. A repetition guard detects identical tool
calls returning identical results and injects corrective feedback from the third
repetition onward. The final submission carries an explicit status, checked
against the committed set in both directions: declaring a match requires at
least one committed pair, and declaring \texttt{no\_match} requires none, so
the agent cannot hedge by abstaining while holding pairs. A task that
ends any other way (an API or transport error, exhausted retries, a timeout, or a
run that stops before submitting) is recorded as \emph{failed}, never coerced into
an empty result. Transient rate-limit errors are retried with deterministic
backoff. Because most chunk pairs share no reuse,
explicit abstention is a first-class outcome.

\section{Expert-Adjudicated Evaluation}
\paragraph{Data.}
Source~$A$ is the \emph{Analects} (20 books, 22{,}919 characters), and source~$B$ is
the \emph{Book of Han} (120 scrolls, 942{,}693 characters), both from a
publicly available punctuated digital edition (Shidian Guji). Their exhaustive
pairing yields 2{,}400 chunk-pair tasks.

\paragraph{Annotation and adjudication.}
Twelve LLMs (below) each processed all 2{,}400 tasks, proposing 6{,}018 candidate
pairs. De-duplicated under a strict fragment-pair criterion (same source pair,
same chunk pair, same fragment texts and same in-chunk offsets), these yield 3{,}489
distinct candidates. Pooling across all twelve models keeps the benchmark
from favoring any single model's proposal distribution. The pool is genuinely a
union, with 68\% of candidates proposed by exactly one model. Manual
enumeration offers no substitute reference. Two philological surveys count
\emph{Analects} quotation in the \emph{Book of Han}; one reports 237, the other
447 \citep{zhang2008liuchuan,wang2014hanshu}. The totals differ by nearly a
factor of two, and both authors expect to have missed cases. Even the larger
falls well below the 954 marked pairs the experts confirm on this source pair
alone. The limit is not effort but sustained attention over a long text, which
is what an exhaustive machine pass supplies.
Two experts independently annotated all
3{,}489 on a purpose-built web interface that shows each candidate as its two
fragments highlighted within fixed context windows: first a validity judgment
(yes/no/uncertain), then, for valid pairs, the five interpretive dimensions,
all forced single-choice. A third expert
adjudicated every item on which the two disagreed, seeing both independent
annotations and the fields in conflict before issuing the final judgment. This yields the adjudicated gold standard: 2{,}533 intertextual pairs,
72.6\% of the pool. Inter-annotator agreement is computed only from the two
independent annotations, before adjudication
\citep{cohen1960kappa,artstein2008intercoder}.

\paragraph{Retrieval baselines: an oracle upper bound.}
The adjudicated set also bounds what any retrieve-then-judge baseline could
recover, since a judge only classifies what retrieval surfaces: 25.8\% of the
2{,}533 pairs share no character trigram between their fragments
(40.8\% share no 4-gram). An edict of Emperor Wu justifies shifting policy priorities by
how Confucius answered each duke differently, ``\zhi{invb}'', naming the sage
yet sharing not one character with the teaching it invokes (``\zhi{inva}'',
Book~13). Ban Gu's
rhapsody in the autobiographical postface compresses the ford encounter of
Book~18 into ``\zhi{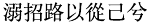}'', recluse and disciple each a single name-character,
unmarked. We give dense retrieval an oracle advantage no deployed pipeline has:
BGE-M3 \citep{chen2024m3} is queried with the expert-identified history-side
fragment itself. Even so, it ranks an overlapping source sentence top-10 for
only 61.0\% of trigram-free pairs (top-1: 33.0\%), against 91.7\% for
verbatim-anchored ones. The edict and the rhapsody rank \#232 and \#470 among
1{,}470 \emph{Analects} sentence units. The stratum's median rank is 4, against
1 for verbatim-anchored pairs. These oracle hit rates cap the recall
of any retrieve-then-judge pipeline exactly where interpretation begins (the
fragment-level counterpart of the chunk-pair bound above). Moreover, retrieval
output carries neither spans nor labels.

\paragraph{Validity: a threshold gap, closed by adjudication.}
On the validity judgment the two experts differ in strictness, not in
criterion. One accepted 55.3\% of candidates, the other 99.4\%. Their
decisions nest perfectly: no candidate was accepted by the stricter expert
and rejected by the lenient one. The disagreement is a one-sided threshold gap
of 1{,}539 candidates, adjudicated item by item, the arbiter siding with the
stricter expert on 59.9\% and accepting 617. Label conflicts also sent 1{,}214
of the both-accepted candidates to the arbiter, who overturned 14 of them to
invalid. The arithmetic closes: $1{,}930-14+617=2{,}533$. Marginals this skewed make chance-corrected
agreement coefficients uninformative \citep{feinstein1990kappa}. The check
that matters is robustness. Rescored against the stricter expert alone, all twelve models'
precisions drop by 9--23 points. However, the ranking (Spearman $\rho=0.94$,
no move over two places) and the deployment choice are unchanged. Absolute
precision inherits the adjudicated threshold, and the comparison does not
depend on it.

\paragraph{A reliability gradient.}
Agreement on the five interpretive dimensions, computed over the 1{,}930
candidates both experts accepted, is not uniform
(Table~\ref{tab:agree}). Dimensions whose evidence sits on the surface of
the text (is the wording identical, is the source named) are settled by
inspection, while \emph{function} and \emph{stance}, which require
reconstructing intent and attitude, admit principled disagreement between
competent experts. Raw agreement is read against chance: \emph{aspect}'s
98.9\% rides a 97\%-content base rate (chance 96.1\%), so its label is
reliable but nearly constant. The gradient bounds what the annotations
support. The distributions of form and source-marking may be discussed with
confidence, while function and stance serve only as exploratory signals. The
disagreement is a property of the task, not a failure of annotation.
Moreover, the models reproduce the gradient where it is informative. Relative
to each model's majority-class baseline, label accuracy lifts by 18--37
points on \emph{form} and 25--43 on \emph{source-marking}, by 12--25 on
\emph{function}, and by at most 2 on \emph{aspect} and 9 on \emph{stance}
(per-model table in the
supplementary material). One contested item makes this concrete: both experts accepted the pair linking the \emph{Analects} phrase
``\zhi{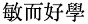}'' (diligent and fond of learning) to the \emph{Book of Han} scene
where a courtier defends the crown prince: ``what is called
talent is: diligent and fond of learning, reviewing the old to know the
new, and such is the crown prince'' (the second phrase itself quotes
\emph{Analects}~2, a two-source composite). Both agreed it is an unmarked
verbatim quotation with positive stance. They split only on \emph{why}: canonical
\emph{support} for the courtier's claim, or \emph{application} of the canon's
standard of talent to a living person (the arbiter ruled application). Nothing
on the textual surface settles this, and 345 items in the pool share this shape:
pair accepted, form and source-marking agreed, function split between support
and application.

\begin{table}[t]
\centering
\small
\resizebox{\columnwidth}{!}{%
\begin{tabular}{lrrr}
\toprule
Dimension & Obs.\ agr. & Chance & Cohen's $\kappa$ \\
\midrule
\icn{quote_gray}\, Form (quotation/paraphrase) & 87.3\% & 56.1\% & 0.71 \\
\icn{layers_gray}\, Aspect (content/structure) & 98.9\% & 96.1\% & 0.70 \\
\icn{tag_gray}\, Source-marking (explicit/marked/unmarked) & 85.9\% & 39.8\% & 0.77 \\
\icn{target_gray}\, Function (background/support/\ldots) & 58.9\% & 39.5\% & 0.32 \\
\icn{thumbs-up_gray}\, Stance (positive/neutral/\ldots) & 77.0\% & 68.4\% & 0.27 \\
\bottomrule
\end{tabular}}
\caption{Inter-annotator agreement on the five interpretive dimensions, over
the 1{,}930 candidates both experts accepted (pre-adjudication). Chance:
expected agreement from the annotators' marginals, the baseline $\kappa$
corrects for. Form and source-marking are reliable; function and stance are
contested; aspect is reliable but near-constant.}
\label{tab:agree}
\end{table}

\begin{figure}[t]
\centering
\includegraphics[width=\columnwidth]{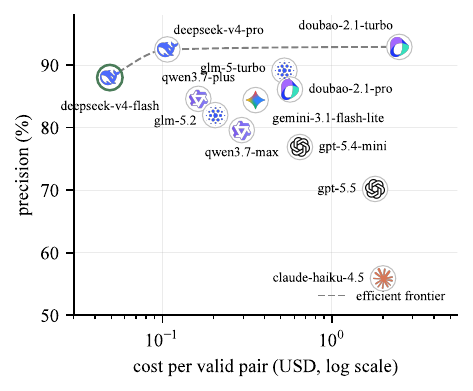}
\caption{Twelve LLMs on the 2{,}533-pair gold standard: precision against cost per
valid pair (log scale). Quality and efficiency vary independently; the selected
deployment model (deepseek-v4-flash) attains the best precision-per-cost
(counts in Table~\ref{tab:models}).}
\label{fig:models}
\end{figure}

\begin{figure}[t]
\centering
\includegraphics[width=\columnwidth]{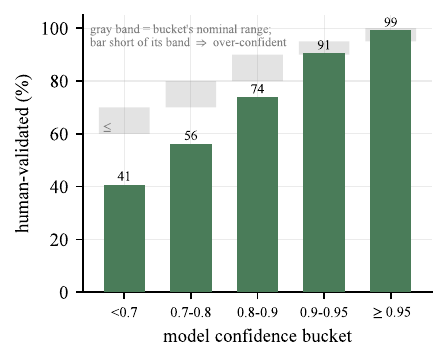}
\caption{Calibration of verbalized confidence, pooled over the twelve models:
share of candidates judged valid per confidence bucket.}
\label{fig:calib}
\end{figure}

\paragraph{Twelve-model study.}
Against the gold standard we score twelve recent LLMs from seven providers
\citep[among them][]{deepseek2026v4,glm2026glm5,openai2026gpt55,anthropic2025haiku45}
(Fig.~\ref{fig:models}; per-model counts in Table~\ref{tab:models}).
Precision ranges from
55.9\% to 92.9\%. Eight of twelve reach at least 82\%, so the strict protocol is
executable for most of the models tested. Quality and efficiency vary independently. At
comparable precision, per-valid-pair cost spans a 51$\times$ range, and the
chosen deployment model attains the highest precision-per-dollar. The
top-precision model is also the slowest (median 521s)
and the most expensive per valid pair. One of its valid pairs buys fifty-one
from the deployment model at 4.9 points lower precision. Meanwhile, the fastest
models (37--39s) span a wide precision range, so precision, cost, and latency
are three separate axes. Verbalized model confidence
\citep{guo2017calibration,tian2023justask}, pooled over the twelve models,
discriminates valid from invalid
candidates (AUC 0.85). Candidates at
confidence $\geq$0.9 are 96.6\% valid (the deployment model alone: AUC 0.84,
96.1\%). However, models are systematically
over-confident at lower scores (Fig.~\ref{fig:calib}). Finally, inter-model agreement is a
strong validity signal. Candidates proposed by $\geq$2 models are 88.5\% valid,
against 65.1\% for single-model ones. Models find different slices of the
intertextual space, which the pooled evaluation aggregates. A single pass
already rivals hand enumeration. Philological surveys of this source
pair counted 237 and 447 quotations by hand
\citep{zhang2008liuchuan,wang2014hanshu}, while each model alone commits 323
to 572 adjudicated-valid pairs.

\begin{table}[t]
\centering
\small
\resizebox{\columnwidth}{!}{%
\begin{tabular}{lrrrrrr}
\toprule
Model & Cand. & Valid & Prec. & Cost (\$) & \$/valid & Lat.\,(s) \\
\midrule
\icn{doubao}\, doubao-2.1-turbo      & 367 & 341 & 92.9\% & 853 & 2.501 & 521 \\
\icn{deepseek}\, deepseek-v4-pro     & 426 & 394 & 92.5\% & 42 & 0.107 & 101 \\
\icn{zhipu}\, glm-5-turbo            & 460 & 410 & 89.1\% & 215 & 0.525 & 126 \\
\icn{deepseek}\, deepseek-v4-flash   & 443 & 390 & 88.0\% & 19 & 0.049 & 131 \\
\icn{doubao}\, doubao-2.1-pro        & 495 & 426 & 86.1\% & 241 & 0.566 & 183 \\
\icn{qwen}\, qwen3.7-plus            & 492 & 416 & 84.6\% & 68 & 0.163 & 178 \\
\icn{gemini}\, gemini-3.1-flash-lite & 384 & 324 & 84.4\% & 115 & 0.355 & 37 \\
\icn{zhipu}\, glm-5.2                & 583 & 478 & 82.0\% & 98 & 0.205 & 162 \\
\icn{qwen}\, qwen3.7-max             & 520 & 414 & 79.6\% & 122 & 0.294 & 194 \\
\icn{openai}\, gpt-5.4-mini          & 455 & 350 & 76.9\% & 227 & 0.648 & 39 \\
\icn{openai}\, gpt-5.5               & 815 & 572 & 70.2\% & 1{,}029 & 1.799 & 38 \\
\icn{claude}\, claude-haiku-4.5      & 578 & 323 & 55.9\% & 647 & 2.003 & 163 \\
\midrule
Total & 6{,}018 & 4{,}838 & & 3{,}676 & & \\
\bottomrule
\end{tabular}}
\caption{Per-model results on the 2{,}400 evaluation tasks: candidates, valid
pairs, precision (validity), total
cost, cost per valid pair, and median latency. Precision is the share of
committed pairs judged valid in expert adjudication.}
\label{tab:models}
\end{table}

\section{Scaling to the Twenty-Four Histories}
We run the selected extractor (deepseek-v4-flash; \citealt{deepseek2026v4}) over the full Twenty-Four
Histories: the \emph{Analects} paired exhaustively with every history scroll,
65{,}380 chunk-pair tasks over $\approx$28.5M characters whose composition
dates span eighteen centuries (91~BCE--1739~CE), with no failed
tasks, yielding 5{,}766 intertextual pairs. Because precision was measured only
on the \emph{Book of Han}, we read the output as a distribution of calibrated model
judgments and state uncertainties explicitly. At the
deployment model's validated precision, roughly one pair in eight is expected to
be spurious. Claims therefore rest on distributional structure rather than raw
totals, and each pair's confidence lets a reader threshold harder.

\begin{figure}[t]
\centering
\includegraphics[width=\columnwidth]{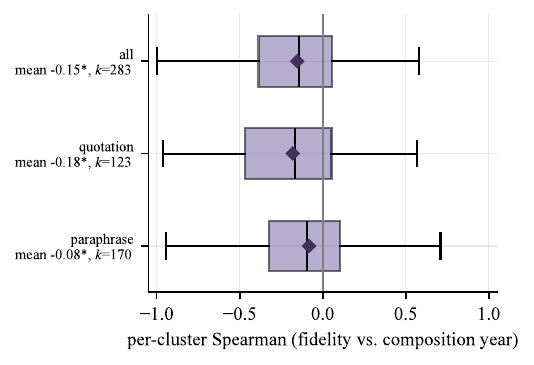}
\caption{Literal fidelity declines with the citing history's composition date,
held constant per passage: distribution of per-cluster Spearman correlations of
fidelity vs.\ year (boxes: interquartile range; diamonds: stratum mean; $^{*}$
permutation $p<0.05$; $k$ qualifying clusters).}
\label{fig:decay}
\end{figure}

\paragraph{A worked example.}
The introduction's abdication charge forms a single merged-fragment cluster of
85 extracted pairs across 22
histories, spanning composition dates from 91~BCE to 1739~CE. Three members show
what the labels add. In the \emph{Book of Han}, the charge is quoted
near-verbatim under the source's name, ``\zhi{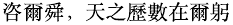}'', labeled
\emph{quotation~/ explicit~/ support} (similarity 0.87): the canon is named to
lend authority. In the same history, an imperial patent appointing a commander
absorbs the charge's wording, ``\zhi{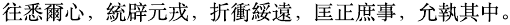}'' (go, devote your whole heart
\ldots\ hold faithfully to the mean), labeled
\emph{quotation~/ unmarked~/ application} (similarity 0.37): investiture
language repurposed as working government prose. In the
\emph{Records of the Three Kingdoms}, the abdication edict transfers the mandate
with ``\zhi{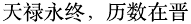}'' (Heaven's favor ends forever; the succession lies with
Jin), labeled \emph{paraphrase~/ unmarked~/ application} (similarity 0.25): the dead
formula of the introduction, now datable and countable. A similarity ranking
orders these three by overlap alone. The labels separate public appeal to
canonical authority, silent administrative repurposing, and formulaic script,
and the cluster assembles the passage's reception history. Within this one
cluster, fidelity already trends downward with composition year, the pattern
quantified across all clusters below (Fig.~\ref{fig:decay}).

\paragraph{The mix of citation shows no detectable drift; volume and fidelity do.}
Across the twenty-four histories we detect no systematic diachronic change in
the interpretive composition of \emph{Analects} citation
(Table~\ref{tab:diachronic}). Comparing
histories composed by 659~CE against those from 945~CE onward, the source-marking
mix is unchanged (unmarked 77.5\% vs.\ 77.6\%), as are the function and stance
mixes. In addition, per-history usage profiles do not diverge with
composition date (Jensen--Shannon divergence uncorrelated with the year gap,
permutation $p$ = .12--.28). The table tracks each dimension's dominant
category, the divergence test the full mix.

\begin{table}[t]
\centering
\small
\resizebox{\columnwidth}{!}{%
\begin{tabular}{llrrrr}
\toprule
Dimension & Dominant category & Early & Late & $\rho$ & $p$ \\
\midrule
\icn{quote_gray}\, Form           & quotation & 40.9\% & 42.4\% & $-0.07$ & .73 \\
\icn{layers_gray}\, Aspect         & content   & 98.1\% & 98.2\% & $-0.28$ & .18 \\
\icn{tag_gray}\, Source-marking & unmarked  & 77.5\% & 77.6\% & $0.11$  & .60 \\
\icn{target_gray}\, Function       & support   & 67.8\% & 70.8\% & $-0.11$ & .61 \\
\icn{thumbs-up_gray}\, Stance         & positive  & 73.8\% & 79.8\% & $0.16$  & .44 \\
\bottomrule
\end{tabular}}
\caption{Diachronic stability of the interpretive composition. Early/Late:
pooled share over histories composed by 659~CE vs.\ from 945~CE onward.
$\rho$, $p$: Spearman correlation of the dominant category's per-history share
with composition year over the 24 histories, permutation $p$ (2{,}000
permutations); no trend reaches significance.}
\label{tab:diachronic}
\end{table} Raw citation \emph{density} does fall
about threefold, but this is a count-level trend obtainable without our
framework, consistent with the received periodization
\citep{pi1959jingxue}, so we rest no claim on it. What the framework
adds is that the \emph{way} the canon is cited shows no systematic change:
its rhetorical form was fixed early and persisted, a stability only annotation
can establish.
This stability does not conflict with the constant rewriting of individual
passages below. A cultural-attraction account expects a stable
group-level distribution produced by convergent transformation, not faithful
copying \citep{buskell2017attractors}. Unmarked
absorption as the durable norm is, in native terms, licensed practice:
the ``compositional'' register, which \citet[Shuolin]{zhangxuecheng_wenshi}
distinguishes from the ``evidential'':
\begin{center}\includegraphics[width=\columnwidth]{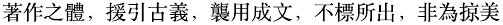}\end{center}
in composition one draws on old meanings and takes over existing text without
marking the source, and this is no plagiarism, a license Zhang grants only
where the author's own argument stands without leaning on the borrowed words.
The norm is older than Zhang's
formulation: Liu Xie's sixth-century poetics already holds that old material,
aptly used, reads ``as if it came from one's own mouth''
\citep[Shilei]{liuxie_wenxin}. The corpus shows that this ideal was, measurably
and durably, the practice.

\paragraph{Marking and fidelity are distinct axes.}
Aligning each history fragment to its \emph{Analects} source (character-level
longest common subsequence) shows that
how literally a citation is quoted does not fix how it is marked. Named citations
are more faithful than unattributed ones (median 0.78 vs.\ 0.38; Kruskal--Wallis
$p<10^{-3}$), and the gap survives controlling for form (0.85 vs.\ 0.50 among
direct quotations), so a similarity score cannot stand in for the marking
label, the point of a reuse typology. Naming a source (``the Master said'') is a
public appeal to canonical authority that holds the wording to a higher standard;
the relation is a regularity, not a law (unmarked citation can be verbatim when
concealment is the motive; \citealt{edelstein2013quote}).

\paragraph{Function tracks fidelity; application is unattributed.}
Invoked as authoritative \emph{support}, the canon is quoted more literally
(mean similarity 0.50); turned to \emph{application}, repurposed into governance
and policy, it is looser (0.42) and 93\% unattributed: the pragmatic use native
scholarship names \emph{tong-jing zhi-yong}, applying the classics to affairs,
where a citation's force is that the sage ``would have approved'' the decision,
not its exact words \citep{yu2021confucian}. Because function is a low-agreement dimension
(Table~\ref{tab:agree}), we report this as exploratory.

\paragraph{Fidelity declines in later-compiled histories.}
Holding a passage fixed (a merged-fragment cluster), later-compiled histories
show lower literal fidelity
(Fig.~\ref{fig:decay}): across 283 clusters spanning $\geq$5 distinct
composition years, the mean Spearman correlation is $-0.15$ (69.3\% negative;
cluster-bootstrap 95\% CI $[-0.19,-0.12]$; permutation $p<0.001$), and the
decline is steeper for direct quotation ($-0.18$) than paraphrase ($-0.08$).
The pattern strengthens in the
confidence $\geq$0.9 subset (172 clusters, $\rho=-0.19$, $p<0.001$). That transmission necessarily alters wording
is, again, a native observation \citep[Shuolin]{zhangxuecheng_wenshi}:
\begin{center}\includegraphics[width=\columnwidth]{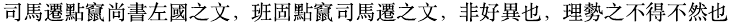}\end{center}
successive historians recast their sources' wording not from a taste for
difference but by
the necessity of circumstance, which is the transformation-as-limiting-case
view of cultural transmission \citep{sperber1985epidemiology}.

\paragraph{Asymmetric reuse fixed segmentation cannot align.}
Reading each chunk in full, rather than matching pre-segmented sentences, also
recovers reuse whose two sides differ in extent, the asymmetric case a fixed
segmentation cannot represent \citep{lau2024mining}. In the \emph{Book of Han}, a
single clause praising Emperor Cheng's bearing, ``\zhi{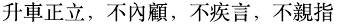}'' (on mounting the
carriage he stood straight; he did not look round, spoke not hastily, pointed at
nothing), draws on two separate passages of \emph{Analects} ``Xiangdang'':
``\zhi{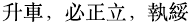}'' (mounting the carriage, he stood straight and grasped the cord) and
``\zhi{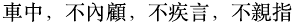}'' (in the carriage he did not look round, spoke not hastily, pointed
at nothing). Both were accepted against the one history clause. A
sentence-to-sentence
ranking must pick one \emph{Analects} sentence or split its score across the two;
the many-to-one relation is expressible only when fragments are localized freely.
It surfaces here as two accepted pairs sharing one history-side span, one of
117 groups in the corpus run where a single history span answers to two or more
disjoint \emph{Analects} passages.

\section{Discussion and Limitations}
Span-grounded records placed under a reuse typology turn detection into
description: a corpus-scale run can now say what kind of reuse occurs, under
what marking, to what rhetorical end. The object of interpretation here is the
textual relation, not the model's reasoning.
Expert agreement bounds the account: form and source-marking are
annotated consistently and carry the load-bearing findings, while function and
stance are contested and reported as exploratory.
First, corpus-scale extraction uses a single model calibrated
on the \emph{Book of Han}. The load-bearing findings persist and strengthen in the
confidence $\geq$0.9 subset (55.3\% of pairs; the deployment model's
$\geq$0.9 candidates are 96.1\% valid on the benchmark),
so they are not carried by lower-confidence pairs. Second, the pool is the union of twelve models'
candidates, as in pooled evaluation \citep{buckley2004incomplete}, so corpus
totals are lower bounds. Third, the model may have absorbed prior scholarly identifications, so we
frame the system as assistive. Fourth, each history enters the diachronic analyses with a single composition
date, though several took decades, and the early/late split follows
received periodization. Contamination
also cuts unevenly: memorized parallels concentrate in famous, marked
cases, while the corpus-scale value lies in unmarked paraphrase, which
concordances do not list.
An anonymized platform (URL withheld) shows every pair in context, with
spans, in clusters tracing reception; the full pair set ships in the
code-and-data supplement as JSONL, SQLite, TEI stand-off XML, and
RDF/Turtle.

The benchmark is a controlled evaluation setting, not a universal
intertextuality benchmark; what travels is the protocol. Little is
specific to the \emph{Analects} or to
Chinese: the task requires only a segmentable source pair and an annotation
schema, and the three principles apply
wherever reuse must be tied to exact spans.
Next: ensemble extraction
and a second adjudicated evaluation on a later history.

\section{Conclusion}
We recast fine-grained intertextuality extraction as a grounded, typology-based
agentic task, built an expert-adjudicated benchmark whose reliability gradient
marks which dimensions annotation can support, and
applied the validated extractor at corpus scale. Across eighteen centuries the
interpretive composition of citation shows no systematic change while the same
passage is
quoted ever less literally. Stability in the aggregate produced by transformation
in the individual case, as a cultural-attraction account expects, and structure
no similarity score can state. The finding is legible only because every pair is a write-time-checked
commitment, so the run is audited rather than trusted.
Comparison once proceeded one book at a time. It now keeps pace with the
corpus.

\section*{Acknowledgments}
This work was supported by the National Key Research and Development Program of
China (``Research on Cross-Context Barrier-Free Interactive Technology and
Equipment for Ethnic Regions,'' No.~2025YFC3309300) and the National Natural
Science Foundation of China (``The Construction of the Knowledge Graph for the
History of Chinese Confucianism,'' No.~72010107003).

\bibliography{references}

\end{document}